# Some Experiments on the influence of Problem Hardness in Morphological Development based Learning of Neural Controllers.

M.Naya-Varela[1], A. Faina[2] and R. J. Duro[1]

[1] Integrated Group for Engineering Research,
CITIC (Centre for Information and Communications Technology Research),
Universidade da Coruña, Spain
{martin.naya, richard}@udc.es
[2] Robotics, Evolution and Art Lab (REAL),
Computer Science Department,
IT University of Copenhagen, Denmark
anfv@itu.dk

**Abstract.** Natural beings undergo a morphological development process of their bodies while they are learning and adapting to the environments they face from infancy to adulthood. In fact, this is the period where the most important learning processes, those that will support learning as adults, will take place. However, in artificial systems, this interaction between morphological development and learning, and its possible advantages, have seldom been considered. In this line, this paper seeks to provide some insights into how morphological development can be harnessed in order to facilitate learning in embodied systems facing tasks or domains that are hard to learn. In particular, here we will concentrate on whether morphological development can really provide any advantage when learning complex tasks and whether its relevance towards learning increases as tasks become harder. To this end, we present the results of some initial experiments on the application of morphological development to learning to walk in three cases, that of a quadruped, a hexapod and that of an octopod. These results seem to confirm that as task learning difficulty increases the application of morphological development to learning becomes more advantageous.

**Keywords:** Cognitive robotics, morphological development, quadrupedal walking, hexapod walking, octopod walking.

## 1. INTRODUCTION.

In the last two decades, the concept of robotic intelligence has been extended to include the morphology of the robot and its environment, as well as their mutual interactions [1]–[3], as relevant components. In fact, this has resulted in a noticeable growth of the field of Artificial Embodied Intelligence (AEI) [4], [5]. AEI postulates that robot intelligence emerges from the interaction between the robot brain, its morphology and the domain it is faced with. Taking this view, as the field of robotics applications addresses ever more complex sequences of environments, it is often the case that the designer cannot contemplate at design time all the possible domains and tasks the robot will have to perform. This has opened an area of research in which the robot, given a basic set of skills, must be able to adapt and learn new skills at run time, as it meets new domains it has not encountered before. In the literature, this is called the open-ended learning problem [6].

The main framework available to the robotics community in order to tackle open-ended learning scenarios in an incremental manner is that of Developmental Robotics (DR) [7]. DR studies different approaches so that robots can autonomously acquire an increasingly complex set of sensorimotor and mental capabilities through the interaction between their bodies and brains in the sequence of domains they encounter in their lifetime [8]. This field is inspired by the development of experience and motor skills in humans, from childhood to adulthood. An example of a very good review on the topic is the one by Asada et al. [7] who concentrate on the development of higher cognitive functions in infants.

The problem with most of the work on DR is that it has generally been focused on Cognitive Developmental Robotics (CDR) [9]. That is, it has dealt with the development of cognition [7], [10], within a fixed pre-designed body and has not introduced the fact that in natural beings the body also undergoes development during the learning period. In other words, it is not only the cognitive module of the system that changes during development but also the morphology of the system. Thus, the question arises of whether this morphological development provides an advantage to the individual for learning in complex open-ended settings and, if so, how it can be used to this end.

The very close relationship between robot morphology, the environment and the task was already highlighted by Pfeifer [11] even to the point that a given morphology may determine the capabilities a system may display in an environment. He concludes that an optimized morphology for a given task simplifies control, makes the task easier and permits improving performance. On the other hand, poorly chosen morphologies lead to inefficient solutions and increase control complexity and computational cost.

Based on these ideas, many authors [12]–[15] have investigated how the close relationship between the physical design of a robot and that of its controller, as a function of the environment or environments and tasks to perform, can be used in order to produce more efficient robot controllers. However, another question that needs to be clearly addressed, and which has been mostly ignored, is how the morphology-control relationship or coupling can be used in developmental processes. In other words, whether there is an advantage in using the development of the body during learning, especially when considering complex behavioral spaces and tasks that are hard to learn and how this should be done.

In natural systems, morphological development is usually taken as encompassing motor development [16], cognitive development [17], and body maturation [18]. Motor development is concerned with the continuous acquisition of motor skills. Cognitive development, on the other hand, operates at a higher level. It deals with the creation of new world representations and in the consolidation and adaptation of those that were previously obtained. Finally, growth and maturation have to do with the actual change of the body. They involve more than an increase in body size and weight and include subtler physical aspects such as increases in muscle tone, bone mass, extensions in the range of motion of the limbs or improving sensorial capabilities, among others.

Notwithstanding the fact that it is usually a continuous process, morphological development can be represented as a series of stages, from the simplest and starting stage, to the final and most complex one where the body has completely developed and has had a chance to adapt to the environment. With this idea in mind, it is important to point out a very fundamental difference between the operation of morphological development and that of cognitive development in the way how they handle the learning process.

In most of the literature on cognitive development, previously learnt units of knowledge are used as scaffolding to learn more complex knowledge nuggets. This is not the case when the body is changing. The cognitive structures created at each stage of morphological development are the starting point for learning the next stage but they are usually "overwritten" by the new learning processes. These structures are modified and adapted to the physical changes that occur during the morphological development process, giving rise to new cognitive systems that are better adapted to the new morphology (e.g. adult walking is not made up of different bits of baby walking and teenager walking). In other words, unlike most cognitive development approaches, the morphological development process is not an incremental construction procedure where one needs to find primitive blocks in order to combine them into more complex structures. It is rather a way to steer a path through a series of solution spaces for the controller that the system interacts with as it body develops and, if chosen wisely, the hypothesis is that it may facilitate learning the final skill and obtaining simplified and less computational costly solutions [19].

This paper is devoted to studying the effects of morphological development as a function of the rate of growth and the hardness of the task. To this end, we have chosen one main task: walking in a straight line. This task will be learnt with and without morphological development by different types of robots with different levels of complexity: a quadruped, a hexapod and an octopod. Learning controllers for walking is easier for a hexapod or octopod than for a quadruped since it has many more options for stable static walking without having to resort to dynamic walking gaits.

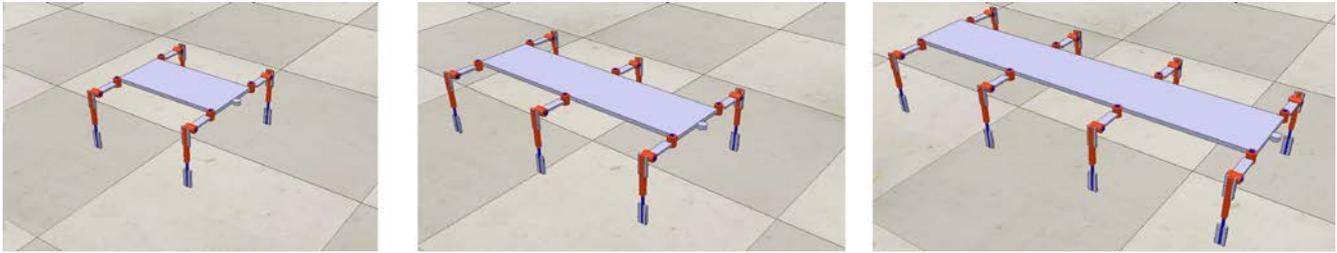

**Fig. 1.** Snapshots of each morphology considered in the experiments in their resting position. Left: quadruped. Middle: hexapod. Right: octopod. Each limb has three solid segments in a chain attached to the base by two actuated revolute joints (red cylinders) and a linear joint (red rectangular cuboid), which is used for the morphological development. The limbs are equidistant from each other at a distance of 0.29cm

The papers we have found related to morphological development for quadrupedal motion do not consider learning per se, but rather, evolution of robots that walk. That is, they are concerned with the use of morphological development during the phylogenetic stage. A representative example is the work by Vujovic [20] who compares the results obtained after applying a strictly evolutionary to those of an EvoDevo sequence (performing development during evolution) to a robot for walking over flat terrain. The length and thickness of the legs were grown during the morphological development stage. His results show that the combination of evolution and development sometimes improves the fitness of the final solution. However, the choice of the way morphological development is carried out is critical. In fact, poorly chosen developmental parameters seem to result in poorer results than considering only evolution.

There are some papers, not many, on morphological development for learning in the case of bipedal [21]–[23] and hexapedal walking [24]. These papers provide examples in which morphological development during learning provides an advantage over just learning with the final morphology, but in other examples the result is just the opposite. That is, their results are inconclusive and most of them indicate that more research is needed in this line.

The objective of this work is to provide some more information and experimental results in order to elucidate whether and how morphological development may lead to a more effective learning process. To this end, the paper is organized as follows: In section 2 we describe the experimental setup we will be using during the experiments. Section 3 is devoted to the presentation of the results of the experiments carried. These results are discussed in section 4 and, finally, we provide some conclusions and future lines of work in section 5.

## 2. EXPERIMENTAL SETUP.

As commented before, the main objective of this paper is to gather data on the influence of morphological development strategies over learning. Out of all the possible morphological development strategies, we will focus on growing limbs and we will concentrate on two aspects in order to construct the experiments. On the one hand, we want to see how problem hardness or learning difficulty affects the influence of morphological development. On the other, for each difficulty, we would like to extract information on what the optimal growth rate would be.

To this end, we have chosen a general experimental structure in which the aim is for the different robots to learn to walk as far as possible. To address how problem hardness affects morphological development, we have assumed that the less possible stable static walking configurations a morphology presents, the more difficult the learning process is and the harder the problem. This is a consequence of the fact that a system with fewer static walking configurations needs to resort to learning dynamic walking strategies, which are much more difficult to learn as most individuals keep falling over. Thus, we have assumed that the more legs a robot has, up to eight, the larger the range of static walking configurations available, making the learning problem easier.

Therefore, we have started with a quadruped robot and added legs. The base morphological design is the quadruped robot, shown in Fig. 1 left. It is made up of a central body of dimensions 30cm*15cm*1cm and 2kg, and 4 limbs, each one with two revolute joints and one prismatic joint. Each limb is composed of 3 segments, all of them present the same size and weight (5cm*2.5cm*0.5cm and 250g). The two revolute joints are actuated. They have a maximum torque of 2.5Nm and they are controlled through a proportional controller (P=0.1). The farthest segments of the robot's legs are joined by the prismatic joint, which is set to a target length for each experiment (maximum force of 50N, P=0.1). The body of the hexapod and octopod is similar to the quadruped robot, but changing the body dimensions to (60cm*15cm*1cm) and (90cm*15cm*1cm) in the case of the hexapod, Fig. 1 middle, and octopod, Fig 1 right, respectively and the number of limbs available, being obviously 6 for the hexapod, and 8 for the octopod. All the prismatic joints of the legs have a maximum stroke of 7.5cm, which means that the length of the legs may vary from 10cm to 17.5cm. In the case of the revolute joint, the maximum motion range available is [-90, 90] degrees.

The controller of the robot is a neural network whose weights and structure are learnt using NEAT [25] specifically the MultiNEAT [26] implementation. It has one input and 8, 12 and 16 outputs respectively, each controlling the actuation of one joint. The input is a sinusoidal function of amplitude 2.0 and frequency 1.0rad/s.

A series of learning experiments using NEAT have been run over different implementations of the robot and environment using the VREP simulator [27] with the ODE physical engine [28] in the CESGA [29] computer cluster. Each NEAT learning run contemplates a population of 50 individuals and is trained for 300 generations. A total of 30 independent runs have been carried out for each experiment with the objective of gathering relevant statistical data. As the controller is obtained using NEAT, the learning strategy is based on a neuroevolutionary process, where the fitness is the distance travelled by the front of the robot. Each individual is tested for 3s with a simulation time step of 50ms and physics engine time step of 5ms.

In order to study how morphological development affects the ability to learn in different morphologies, we have carried out two kinds of experiments for each morphological configuration:

**Reference experiment.** This experiment is run with a fixed morphology (the same as the final morphology for the rest of the experiments) from the beginning to the end. The robot starts at generation 0 with the maximum length of the legs and the neuro-evolutionary algorithm seeks a neural network-based controller to achieve maximum displacement.

**Leg growth experiments.** The robot morphology starts with the shorter version of the legs. That is, at the beginning the prismatic joint is fully contracted, its extension is 0cm, and the length of the legs is thus 10cm. The leg length is grown linearly for a number of generations until it reaches the maximum length of 17.5cms. This growth takes place in a set number of generations for each experiment. That is, the final morphology is reached at generation 20, 40, 60, 80, 100 and 120 depending on the experiment. This permits studying the relevance of the growth rate with regards to performance. In a way, we try to simulate the way knowledge is acquired by biological entities: their limbs grow until a certain age, and then learning continues with a fixed morphology.

The results obtained from each of the experiments have allowed us to compare the differences that exist between the methods applied to each morphology, and the differences in results that occur between different designs.

## 3. RESULTS.

The results of the training process for each case can be observed in Fig. 2 and Fig. 3. Figure 2 shows the results obtained after the learning process through neuroevolution, in the case of no-development and growth. It displays the median of the best fitness obtained for the 30 independent runs at each generation for 3 of the growth rates and the reference. The shaded areas in the graph represent the areas between percentiles 75 and 25 for each experiment. Although we have carried out all of the experiments mentioned in section 2, for the sake of clarity in this graph, we only represent the experiments where the final morphology is reached at generations 20 and 120, as they are the fastest and slowest growth rates respectively and the case for which the best development result was obtained for each morphology. For the quadruped, this corresponds to reaching the final morphology in generation 60, Fig 2. top left, and for the octopod ingeneration 100, Fig. 2 bottom. In the case of the hexapod, Fig 2. top right, the most statistically relevant results were reached in the experiment where the legs grow up to generation 20. Consequently, we also include in this graph the case for which the maximum median was achieved (growth up to generation 100).

Two important points can be observed in these graphs:

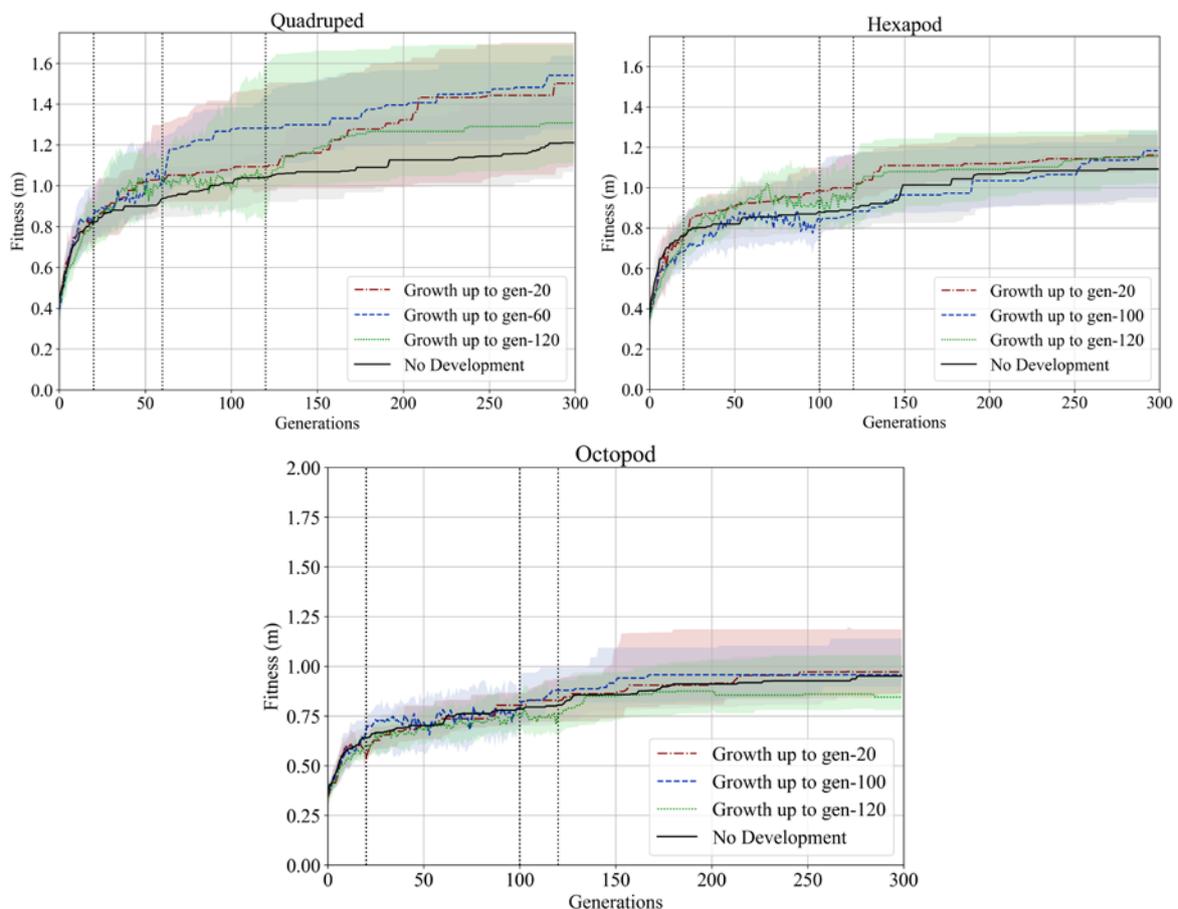

**Fig. 2.** Statistical representation of the neuroevolution of 30 independent experiments for each morphological configuration. All three images show results for the no-development case (black line), growth up to generation 20 (red dashed line) and 120 (green dotted line). Furthermore, the best growth ratio is shown for each morphology. Top left: Quadruped topology, for which the best results were obtained for growth up to generation 60. Top right: Hexapod topology and highest median of growth up to generation100. Bottom: Octopod topology and best case of growth up to generation100.

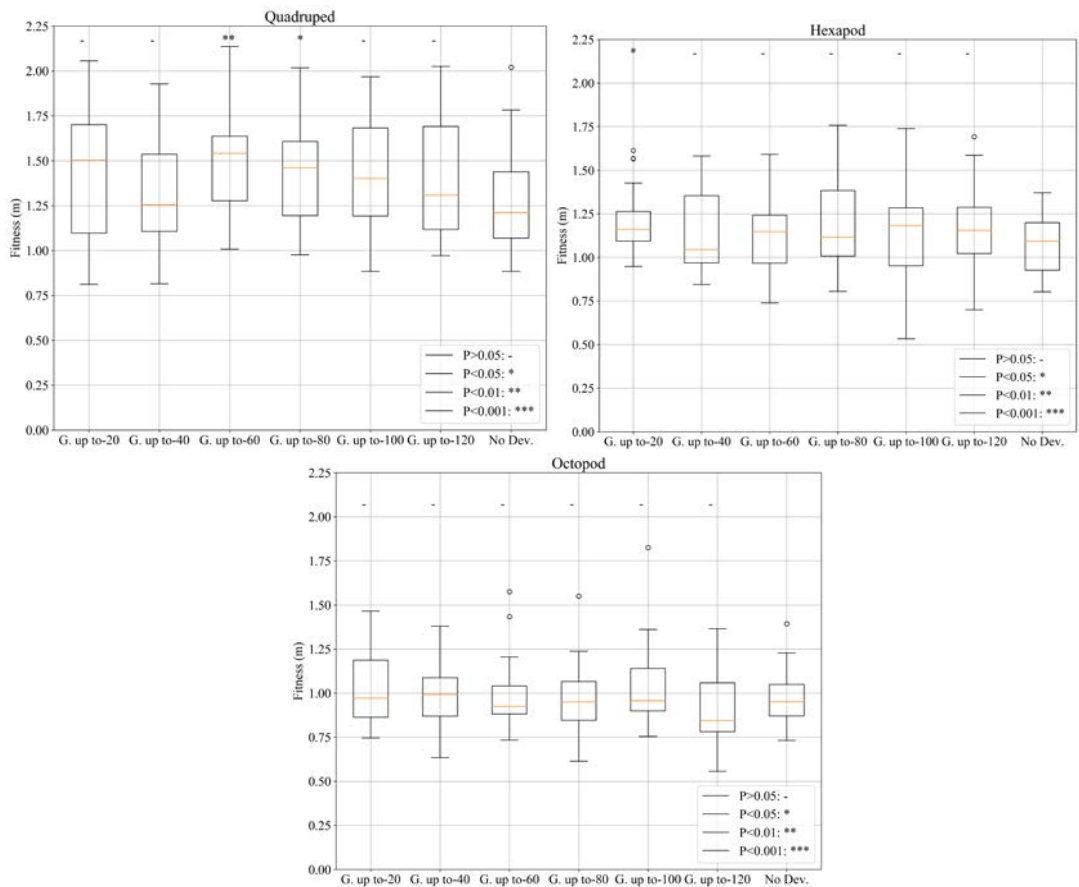

**Fig. 3.** Statistical representation of the performance obtained at the end of the neuroevolutionary process from the 30 independent experiments for each morphology. Top left: Statistical results for the quadruped morphology. Top right: Statistical representation for the hexapod morphology. Bottom: Statistical representation for the octopod morphology. Asterisks and lines refer to the statistical difference between the samples represented. Their corresponding numerical values are shown in the legend of the figure.

**Growth performance variation.** It can be easily observed that the morphological development mechanism based on leg growth provides better results than when there is no development in the case of the quadruped. Better results are also observed for one of the growth rates of the hexapod. In the case of the octopod, there is no significant improvement between using morphological development through growth and no-development.

This analysis is supported by the results presented in Fig. 3. Each boxplot represents the median and the 75 and 25 quartile in the last generation for 30 independent runs of each of the different types of experiments. The whiskers are extended to 1.5 of the interquartile range (IQR). Single points represent values that are out of the IQR. All developmental samples are compared to the no-development case. The statistical analysis has been carried out using the two-tailed Mann-Whitney test. We want to test whether the null hypothesis (if both compared samples are equal) is true. We consider a p-value of 0.05 as the significant value for accepting or rejecting the null hypothesis.

For the quadruped, the most relevant result is obtained for growth up to generation 60, leading to a p-value of 0.00238, which means a rejection of the null hypothesis. That is, in this case morphological development clearly improves learning with respect to no-development. Growth experiments up to generation 80 present a p-value of 0.01988 showing that development offers statistically relevant better results than no development. For the rest of development ratios, they display worse performance and they are not statistically relevant. For the hexapod, although the highest median is obtained in the growth experiment up to generation 100, the most relevant result is obtained for growth up to generation 20, with a p-value of 0.01221. This statistical significance indicates that both samples can be considered different, and therefore allows us to confirm that this morphological development mechanism favors learning in this case. None of the rest of the experiments, can be considered better than the no-development case. For the octopod the best p-value, 0.596, is obtained in the experiment for growth up to generation 100, although there are no statistically relevant results showing any difference, and therefore, morphological development cannot be considered better than no-development.

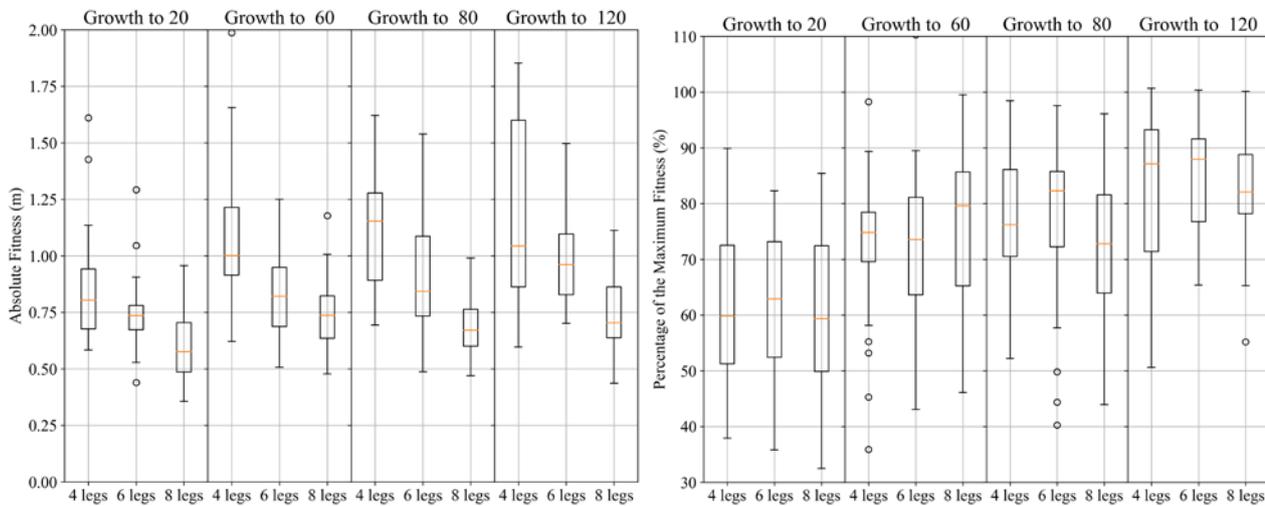

**Fig. 4.** Left: Statistical representation of the performance obtained at the end of each developmental process for the 30 independent experiments for each morphology. Right: Statistical representation of the relative performance, as a percentage of the maximum fitness achieved at the end of each learning process (generation 300).

**Morphological fitness variation.** If we take as a reference the median of each of the experiments of both development and no-development for each morphology, the slope of the learning curves, as well as the maximum fitness achieved at the end of the learning process varies with each morphology. Fig. 4 left shows the statistical results of the absolute fitness value of different growth experiments up to generation 20 and 120, as they are the fastest and slowest growth rates respectively, and those which grow up to generation 60 and 80, as intermediate growth rates. If we compare the three morphologies, in the four cases for each growth ratio the highest absolute values are obtained, in decreasing order, for the quadruped, the hexapod and the octopod. These results are in accordance with those displayed in Figure 3, where it was shown that at the end of the neuroevolutionary process, the best results in absolute value were found for the quadruped, and the worst for the octopod. In Fig. 4 left we want to highlight how the dispersion of fitness varies as growth speed decreases and morphological complexity increases. This dispersion increases as the growth rate decreases and decreases with the complexity of the controller and the constraints on the morphology, being the quadruped the one that exhibits largest dispersion and the octopod the smallest with the decrease in growth speed. This implies a direct relationship between the morphology and the best possible performance of the task. This can be explained by the fact that morphologies with more legs are more constrained in terms of how much the body can twist. This implies that morphologies with less legs are able to combine body twist and leg reach in order to provide longer strides, thus making them able to walk farther in the same amount of time.

However, if we change these results into relative values, understanding relative values as the absolute values shown in Fig. 4 left, divided by the maximum fitness value reached at the end of the learning process for each experiment expressed as a percentage, the graph changes remarkably (Fig. 4 right). Although the statistical dispersion varies significantly from one growth ratio to another, it can be seen how the values of the median are quite similar, being 9.5274% maximum difference, for growth up to generation 80. Fig 4 left and Fig 4 right, seem to lead to the conclusion that the maximum fitness achievable and thus, the fitness increase margin is dependent on the constraints the morphology imposes and not on the morphological development strategies.

Considering this, from Fig. 2 and Fig. 3, we may assume that morphological development offers an advantage in those cases in which the solution space available to the learning process changes more drastically as the morphology changes. In other words, when it is hard for the grown individual to find optimal solutions due to the very small attraction basin of optimal, or even reasonable solutions, leading to a high dispersion or very poor results at the end of the neuroevolution in the case of no development. In these cases, an appropriately selected morphological development strategy that seeks to expand the attraction basins of the optimal solutions when the individual is "young" and progressively lead to the final configuration helps to lead the learning process towards the most promising areas of solution space avoiding getting stuck in local minima and, thus, biasing the search towards the optimal solutions. That is why development offers better results with the quadruped: its space of optimal solutions is narrower than in the case of the hexapod and the octopod and growth helps to move towards optimal solutions. In the hexapod and, especially in the case of the octopod, the solution spaces at the beginning and end of the growth process are very similar due to the constraints imposed by the morphology and, thus, morphological development offers less of an advantage.

## 4. DISCUSSION.

While our aim in this paper was to study how the hardness of the task to learn affects the benefits of morphological development, there are a few limitations in our methodology.

Each type of morphology shows different absolute fitness values. This is caused by the fact that the morphologies have different physical features and because the learning can be more complex in some morphologies than in others. Regarding the physical features, the controllers have 8 outputs for the quadruped, 12 for the hexapod and 16 for the octopod and the central body is heavier and bigger in morphologies with more limbs. Regarding the difficulty of learning, there are different aspects to consider. On one hand, due to the number of variables involved in the problem, we presume that it is more complicated to efficiently control 16 outputs than 8. On the other, the octopod is very stable and its twist motion very constrained. Therefore, there are a lot of controllers that can effectively move the robot, even in its final morphology. On the contrary, the quadruped has fewer actuators to control but it is very unstable and much less constrained in its overall motion. This means that a lot of controllers will present very poor fitness values even if they are close to the optimum, creating a very deceptive search space, which makes learning more difficult. However, the controllers that are successful will present much higher fitness values as dynamic gaits are very efficient.

We consider that the hardness of learning decreases when adding extra limbs that give stability and constraining the motion of the robot. Therefore, learning the controllers in the grown up octopod is easier than in the grown up quadruped, because almost any controller allows it to move from its initial position without falling over or damaging itself. Learning is simple as there are multiple solutions that perform well and there is a gradient that guides the learning.

In the quadruped, we hypothesize that morphological development leads to an improvement of the performance during neuroevolution because, when the robot is still an infant, it provides the necessary stability and dynamic control that expands the attraction basing of the optimal solutions and allows controllers to explore the space of possible solutions biasing the solutions towards the optimum, moving it away from the stagnation in local minima that happens in the case of no-development.

The effect of morphological development seems to decrease with how easy to learn is a task for the grown up individual. If the task is easier to learn, there is no effect (octopod). If the task is hard to learn, noticeable improvement during learning can be appreciated (quadruped). The effects are visible (two experiments display statistical significance), but the correct growth rate must be selected.

These results show the need to study in greater depth the underlying mechanisms of growth in morphological development applied to robotics, in order to determine which cases can be classified as hard to learn, in order to be sensitive to the application of morphological development.

## 5. CONCLUSION AND FUTURE WORK.

In this paper we have shown that ontogenetic morphological development for quadruped, hexapod and octopod morphologies, while learning a locomotion task under some conditions, may help to find better solutions when compared to a learning process that does not follow any kind of development. In our robots, morphological growth of the limbs can help to learn better controllers if the growth rate is chosen correctly. However, we have also found that an incorrect synergy between morphology, the hardness of the task and development could render this approach irrelevant.

For our studies, it seems that there is a relationship between the utility of applying morphological development and the hardness of the task to be accomplished by the grown up individual: if the task is too easy to learn, even though it may display a high level of complexity, morphological development does not represent any advantage during learning. However, if the task is hard to learn even if it is simple, morphological development is a good methodology to improve learning outcomes.

Notwithstanding the previous comments, much more work is needed over many other morphologies and tasks to really be able to provide effective engineering indications on how to apply growth as a morphological development tool to aid learning. We are currently extending the range of morphologies and cases in order to get a better grasp of the scope of the approach.

### ACKNOWLEDGMENT

This work has been partially funded by the Ministerio de Ciencia, Innovación y Universidades of Spain/FEDER (grant RTI2018-101114-B-I00), Xunta de Galicia and FEDER (grant ED431C 2017/12) and M. Naya-Varela is very grateful for the support of the UDC-Inditex 2019 grant for international mobility. We also want to thank CESGA (Centro de Supercomputación de Galicia. https://www.cesga.es/) for the possibility of using its resources.